\documentclass[10pt,twocolumn,letterpaper]{article}

\usepackage[pagenumbers]{cvpr}

\newcommand{\PAR}[1]{\vspace{0.1cm}\noindent{\bf #1} }

\definecolor{cvprblue}{rgb}{0.21,0.49,0.74}
\usepackage[breaklinks=true,colorlinks=true,linkcolor=blue,citecolor=blue,urlcolor=blue,bookmarks=true,bookmarksnumbered=true]{hyperref}

\newcommand{\secondbg}[1]{\cellcolor{gray!15}{\underline{#1}}}

\usepackage[table]{xcolor}
\definecolor{BestBlue}{RGB}{0,82,204}

\newcommand{\bestbg}[1]{\cellcolor{BestBlue!12}{\textbf{#1}}}

\newcommand{\worstbg}[1]{\cellcolor{orange!20}{\textbf{#1}}}

\usepackage{graphicx}

\usepackage{amsmath,amssymb}
\DeclareMathOperator*{\argmax}{arg\,max}

\usepackage{bm}
\usepackage{etoc}

\def\paperID{}
\def\confName{CVPR}
\def\confYear{2026}

\title{RoboRouter: Training-Free Policy Routing for Robotic Manipulation}

\author{
Yiteng Chen\textsuperscript{1,*},
Zhe Cao\textsuperscript{2,*},
Hongjia Ren\textsuperscript{3,*},
Chenjie Yang\textsuperscript{4,*},
Wenbo Li\textsuperscript{1,*},
Shiyi Wang\textsuperscript{1},\\
Yemin Wang\textsuperscript{5},
Li Zhang\textsuperscript{6},
Yanming Shao\textsuperscript{7},
Zhenjun Zhao\textsuperscript{8},
Huiping Zhuang\textsuperscript{1},
Qingyao Wu\textsuperscript{1}\\[0.5em]
\textsuperscript{1}South China University of Technology \quad
\textsuperscript{2}Nanjing University \quad
\textsuperscript{3}University of New South Wales \\
\textsuperscript{4}Southwest Jiaotong University \quad
\textsuperscript{5}Xiamen University \quad
\textsuperscript{6}The Hong Kong Polytechnic University \\
\textsuperscript{7}ShanghaiTech University \quad
\textsuperscript{8}University of Zaragoza
}

\begin{document}
\maketitle

\renewcommand{\thefootnote}{\fnsymbol{footnote}}
\footnotetext[1]{Equal contribution.}
\renewcommand{\thefootnote}{\arabic{footnote}}

\begin{abstract}
Research on robotic manipulation has developed a diverse set of policy paradigms, including vision-language-action (VLA) models, vision-action (VA) policies, and code-based compositional approaches. Concrete policies typically attain high success rates on specific task distributions, but limited generalization beyond it. Rather than proposing another monolithic policy, we propose to leverage the complementary strengths of existing approaches through intelligent policy routing. We introduce \textbf{RoboRouter}, a training-free framework that maintains a pool of heterogeneous policies and learns to select the best-performing policy for each task through accumulated execution experience. Given a new task, \textbf{RoboRouter} constructs a semantic task representation, retrieves historical records of similar tasks, predicts the optimal policy choice without requiring trial-and-error, and incorporates structured feedback to refine subsequent routing decisions. Integrating a new policy into the system requires only a lightweight evaluation and does not incur training overhead. Across simulation benchmark and real-world evaluations, \textbf{RoboRouter} consistently outperforms individual policies, improving the average success rate by more than 3\% in simulation and 13\% in real-world settings, while preserving execution efficiency. Our results demonstrate that intelligent routing across heterogeneous, off-the-shelf policies provides a practical and scalable pathway toward building more capable robotic systems.
\end{abstract}    
\section{Introduction}
\label{sec:intro}

Robotic manipulation in diverse, unstructured environments remains one of the grand challenges in embodied intelligence. Recent years have witnessed remarkable progress through multiple research directions: diffusion-based visuomotor policies~\cite{chi2025diffusion,ze20243d,xue2025reactive,wang2024rise,su2025dense,zhang2025chain} achieve stable action generation; vision-language-action (VLA) models~\cite{kim2024openvla,liu2024rdt,shukor2025smolvla,li2025cogvla,ni2025swiftvla,yang2026abot,zeng2025janusvln,chen2025sifthinker,zhang2025vision,li2025semanticvla,bu2025univla,zhao2025bias,li2025bridgevla} leverage Internet-scale data for generalist capabilities; and code-based composition methods~\cite{huang2023voxposer,huang2024rekep,pan2025omnimanip,fang2026uncertainty} enable flexible high-level task planning. Each paradigm demonstrates compelling strengths in specific scenarios, yet each also faces fundamental limitations.

VLA models perform well within their training distribution and become particularly strong after reinforcement learning post-training~\cite{li2025simplevla}, but suffer significant performance degradation on out-of-distribution tasks~\cite{zhou2025libero}. Vision-action (VA) models trained for specific tasks achieve strong performance on their target domains but transfer poorly to novel scenarios. Interface-based composition approaches excel at high-level reasoning but struggle to learn fine-grained affordance and contact-rich manipulation from demonstrations. Consequently, despite continuous advances, no single approach dominates across diverse manipulation challenges.

This landscape presents both a challenge and an opportunity. Robotics community~\cite{wolf2019huggingface,lhoest2021datasets} has developed a rich ecosystem of diverse, off-the-shelf policies, each capturing different aspects of manipulation competence, developed by different research groups on distinct datasets and with varied training paradigms. Our empirical analysis reveals a key insight: \textit{while individual policies exhibit inconsistent performance across tasks, strategically combining multiple policies achieves substantially higher coverage and success rates}. This observation motivates a paradigm shift from building monolithic generalist policies to intelligently routing tasks among specialized experts.

We explore a natural question: \textit{can we design a system that automatically selects the most suitable policy for each task?} Such a system must satisfy several practical requirements. First, it should accommodate new policies without expensive retraining, given the rapid pace of progress in robot learning. Second, it must make routing decisions efficiently, without executing all candidate policies. Third, it should improve continuously through online experience. We present {\it \textbf{RoboRouter}}, a training-free routing framework that addresses these requirements through a simple yet effective design.

RoboRouter maintains a pool of heterogeneous manipulation policies alongside a database of historical task executions. For each new task, the system constructs a multimodal representation combining language instructions, visual observations, and task metadata. It retrieves historical executions of similar tasks from the database, reasons about each policy's likelihood of success, and executes the selected policy directly. After execution, a evaluator analyzes the execution video and generates structured feedback, which is stored in the database to inform future routing decisions. This creates a closed-loop system that continuously refines its understanding of policy capabilities. Integrating a new policy requires only lightweight evaluation on a small task set, with no training.

We evaluate RoboRouter across simulation benchmark~\cite{chen2025robotwin} and real-world robot platforms. Our experiments show that routing among existing policies consistently outperforms the best individual policy, improving average success rate by more than 3\% in simulation and over 13\% on physical robots. Ablation studies further validate the contribution of each system component.

In summary, our \textbf{contributions} are as follows:
\begin{itemize}
    \item We identify the policy routing problem for robotic manipulation and describe a basic formulation in which a router chooses among heterogeneous policies based on a task representation.
    \item We propose RoboRouter, a training-free routing framework that selects policies through retrieval-augmented reasoning and improves continuously via execution feedback, requiring minimal overhead to integrate new policies.
    \item We conduct extensive experiments on simulation benchmark and real-world platform, achieving state-of-the-art performance and validating design choices through comprehensive ablations. We will release our framework to facilitate future research.
\end{itemize}

\begin{figure}
  \centering
  \includegraphics[width=\linewidth]{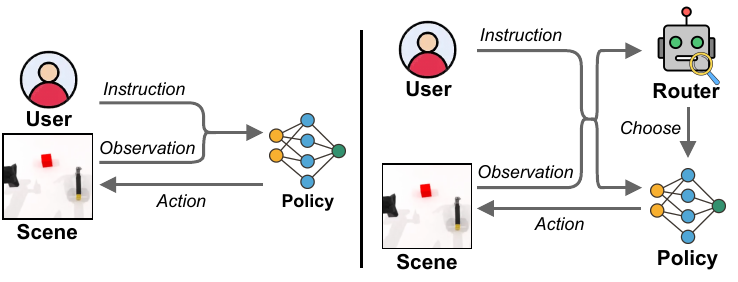}
  \caption{Illustration of conventional manipulation pipeline (left) versus the proposed routing framework (right). Conventional pipeline maps observations and language instructions to actions for all tasks. Our framework augments this setup with a router that selects a suitable policy from a pool of candidates for each episode.}
  \label{fig:pipeline_com}
\end{figure}
\section{Related Work}
\label{sec:relatedwork}

\subsection{Robotic Manipulation Policies}

In recent years, robot manipulation policies have seen notable progress. VA models directly learning vision-to-action mapping, using mechanisms like diffusion generation to deliver stable control~\cite{chi2025diffusion}. Subsequent work extends this paradigm toward improved 3D perception~\cite{fang2025airexo,wang2024rise,zhao2026advances,ze20243d,wang2026radar}, efficient inference~\cite{su2025dense,wu2025device}, and other aspects of visuomotor policy design~\cite{xia2025cage,su2025motion,zhang2025chain,chen2025g3flow}. VLA models~\cite{liu2024rdt,kim2024openvla,black2024pi_0,li2024cogact,bu2025univla,li2025bridgevla,dong2026visnec,shukor2025smolvla,xu2025affordance,xiaoreversible,zhoucomem} further integrate natural language, enabling robots to understand diverse instructions via large-scale multimodal pre-training; however, they depend on large-scale datasets and substantial computational resources, and suffer performance degradation in out-of-distribution tasks. Some works~\cite{liang2023code,huang2023voxposer,huang2024rekep,li2025identify} uses vision-language models (VLMs) as planners that generate code to invoke low-level interfaces, achieving strong task-level generalization but struggling to learn fine-grained affordances and complex strategies directly from demonstrations. These strengths and limitations across paradigms motivate us to integration heterogeneous policies for flexible task transfer without retraining.

\subsection{LLM-based Agents}

Large language models (LLMs) have made substantial progress in reasoning~\cite{wei2022chain,zeng2026halluguard,yu2025mquant,kojima2022large,chen2025think,fang2026reasoning,lin2025plan,chen2025visrl,zhang2025assessing,zhang2024exploring,cai2025does,hu2026codes,li2026vlaattc,li2026sentinelvla,yu2026mathagent,zhou2025gsq,li2026draft,zhou2026mdk12,zhou2025foodsky,zhang2025bifair,gong2024uknow,wang2025reasoning}. Building on these advances, LLM-based agents~\cite{wang2024survey,zhou2023agents,hong2023metagpt,yao2022react,zhang2025rearank,xiang2025promptsculptor} have attracted growing attention for their ability to follow user instructions, invoke tools, and solve complex tasks. Some prior work has explored constructing task-specific agent systems~\cite{wang2023voyager,huang2026learning,tang2026agent,xu2024fakeshield,zhou2023webarena,wu2024autogen,zhang2024generative,zhangpi,ren2024embodied,wei2025copeft} tailored to particular application domains, as well as equipping agents with memory, for example through external vector-database–based memory modules~\cite{gao2023retrieval,lewis2020retrieval,guu2020retrieval} or by explicitly training agents to manage and update their own memories~\cite{yu2025memagent,wang2025mem}. RoboRouter follows this line of research with a multi-agent architecture that leverages historical execution records to route incoming robotic manipulation tasks to the most suitable control policy.

\subsection{Model Routing}

\begin{figure*}[t]
  \centering
  \includegraphics[width=\textwidth]{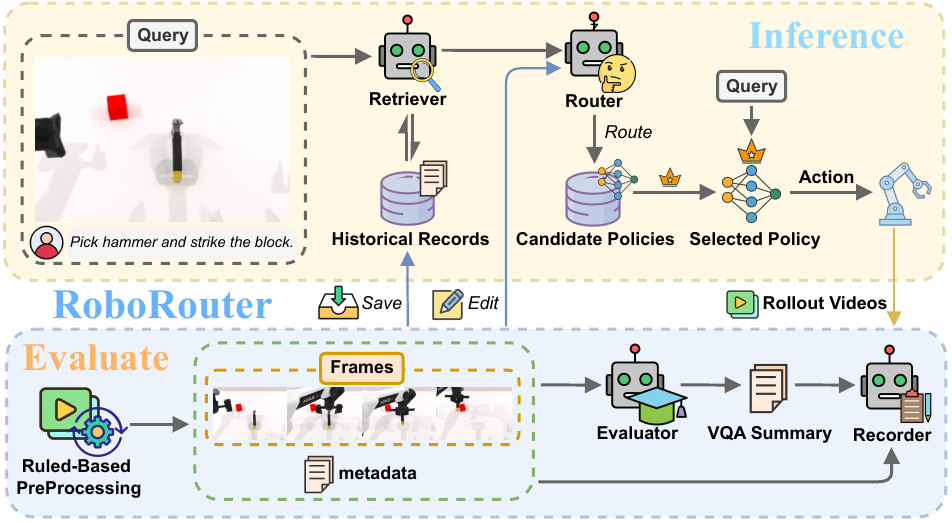}
  \caption{Overview of the RoboRouter architecture, composed of four collaborating agents. The inference pipeline (top) uses a Retriever to query historical execution records and a Router to select a suitable policy from the candidate pool for the current episode. The evaluation pipeline (bottom) consumes rollout videos and metadata with an Evaluator and a Recorder to obtain structured feedback. This feedback is used both to register new policies and to continuously refine routing decisions through online updates.}
  \label{fig:overview}
  \vspace{-0.5em}
\end{figure*}

Model routing has emerged as a key mechanism for modular intelligence. Some approaches use routing to balance task performance and computational overhead~\cite{jitkrittum2025universal,feng2024graphrouter,ding2025best}. Others focus on dispatching user queries to specialized expert models~\cite{zhang2023model,zhang2025capability,zhuang2024embedllm}, improving overall performance. There are also training-free framework~\cite{zhao2024eagle} that implement routing without additional fine-tuning. However, most of these methods are developed in the context of LLMs and have not yet been extended to robotic manipulation.
\section{Method}
\label{sec:method}

RoboRouter is a multi-agent policy-routing framework for robotic manipulation. 
We first introduce the framework’s four agents and other core components, and then detail the inference pipeline, the onboarding pipeline for integrating new policies into the pool, and the online feedback pipeline. An overview of the framework is shown in \cref{fig:overview}

\subsection{Problem Formulation}

\begin{figure*}[t]
  \centering
  \includegraphics[width=\textwidth]{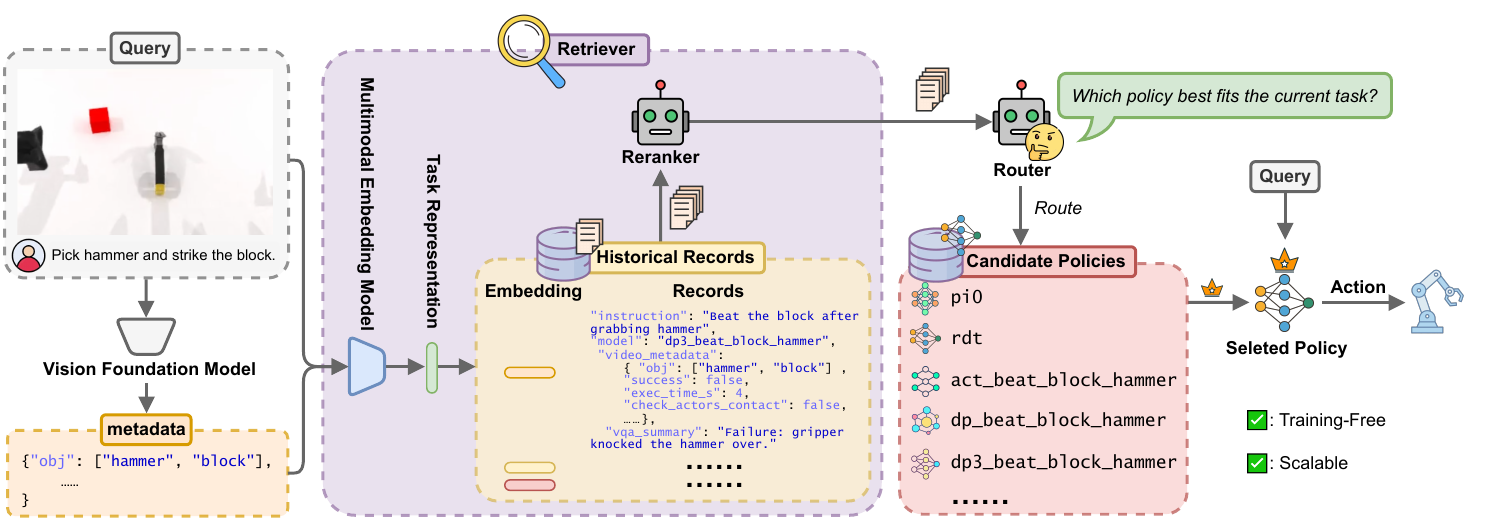}
  \caption{Inference pipeline of RoboRouter. The system first converts instruction and observation into multimodal task representation, together with lightweight metadata from a vision foundation model. Retriever search historical records and a reranker reorders the retrieved records, and Router then selects the policy that best fits the current task from policy pool.}
  \label{fig:infer}
  \vspace{-0.5em}
\end{figure*}

We study embodied manipulation with policy routing, this is a simple yet novel problem in robotic manipulation.
At each discrete time $t=0,1,\ldots$, the system receives a visual observation $o_t$
and a natural-language instruction $l$.
Let $\mathcal{P}=\{\pi^{(m)}\}_{m=1}^{M}$ denote the pool of candidate policies,
where each $\pi^{(m)}$ maps $(o_t,l)$ to a distribution over actions $a_t$.
For routing, we compute a multimodal task representation $z=g(o_t,l)$ and assign each policy a performance score $S_{\pi}(z)$.
The router selects the policy in~\cref{eq:router} and then issues actions as in~\cref{eq:control}.

\begin{equation}
\pi^{\star}(o_t,l)=\argmax_{\pi\in\mathcal{P}} S_{\pi}\!\big(g(o_t,l)\big).
\label{eq:router}
\end{equation}

\begin{equation}
a_t \mid o_t,l \sim \pi^{\star}.
\label{eq:control}
\end{equation}

\subsection{Multi-Agent Framework for Policy Routing}

RoboRouter is a policy-routing framework that selects the most suitable policy for each incoming robotic manipulation task. It comprises four agents built on LLMs/VLMs~\cite{hurst2024gpt,achiam2023gpt,team2024gemini,li2025hybrid}: Retriever, Router, Evaluator, and Recorder.

\label{retriever}
\PAR{Retriever.}
Unlike LLM text Question Answering (QA), robotic manipulation tasks are multimodal, combining instruction text, visual inputs, and other signals; many current policies (VA and VLA) are highly sensitive to visual distribution shift (camera pose, background, object appearance, etc.)~\cite{zhou2025libero}. A central question is how to build a discriminative task representation for robotic manipulation that captures as many policy performance-critical factors as possible. 
We propose a simple and effective multimodal representation method for manipulation tasks. Specifically, pretrained visual foundation models~\cite{ravi2024sam} is first used to extract metadata from task images, including task-involved objects and other task-relevant information. Then, a multimodal embedding model encodes the task instruction, image, and metadata to obtain a multimodal embedding, which serves as the task representation. A database is constructed to store historical execution records of various policies on different tasks, where each entry is indexed by the corresponding task representation. During retrieval, the task representation of the current manipulation task is first constructed. Using cosine similarity, retrieves from the history database the execution records of the top-k tasks most similar to the current task, which are subsequently re-ranked by a reranker.

\PAR{Router.}
Router is an LLM-based agent. During system inference (\cref{fig:infer}), it receives historical execution records returned by Retriever for tasks similar to current task; these records cover performance across different policies. Router estimates which policy in the pool is most likely to succeed on current task, outputs a policy-selection decision, and invokes the chosen policy to execute current task. It also maintains a routing context that concisely summarizes each policy’s overall performance across tasks.

\PAR{Evaluator.}
Evaluator is a VLM-based agent that processes each execution and produces structured evidence for downstream routing decisions. Compared with naive success-rate counting, which requires many repeated trials to gather enough evidence and is inefficient, Evaluator simulates a human researcher, observing and analyzing a single run more richly. A lightweight preprocessing step converts the execution video into frames and applies task-specific Python utilities to compute rule-based metrics (success flag, elapsed time, tool–object distance, contact/alignment, etc.), yielding metadata that augments visual frames because occlusion between end-effector and objects may hides task-relevant details. Given task instruction, video frames, and metadata, Evaluator outputs a natural-language Visual Question Answering (VQA) summary: for success, it briefly summarizes the execution; for failure, it describes the specific failure behavior. Evaluator operates in both new-policy onboarding and online feedback pipelines; all outputs are passed to Recorder to update database entries and router context. Overall, Evaluator acts as an information compressor, distilling execution signals, including video streams, into compact, decision-support evidence for routing while avoiding excessive storage cost. Further details and ablations appear in \cref{sec:Experiments}.

\PAR{Recorder.}
Recorder writes each execution result back to system memory, which conceptually consists of two layers: Database and Router Context. Recorder runs on two pipelines: in new-policy onboarding pipeline, it records the performance of newly introduced policies on selected tasks, enabling policy integration; in the online feedback pipeline, it logs outcomes of online executions to support continual feedback learning. Based on the metadata and video summary produced by the Evaluator, Recorder constructs new record entries and incrementally updates the Database using multimodal task indices (see Retriever). Meanwhile, updates to Router Context follow a structured incremental scheme~\cite{zhang2025agentic}, the context is represented as a set of itemized bullet points rather than a monolithic prompt. Recorder appends new entries with unique identifiers or performs in-place updates on existing items, then merges duplicate items using semantic embedding comparisons to reducing redundancy. This mechanism eliminates the computational overhead and detail erosion caused by full context rewrites, allowing Router Context to retain long-term memory while expanding stably, thereby supporting scalable and continuously policy routing frameworks.

\subsection{Inference Pipeline}

In the inference pipeline, we first construct a task representation for the current robotic manipulation task; Retriever retrieves historical execution records of similar tasks. Next, a Router, conditioned on its context and these records, selects from the policy pool the policy with the highest expected success rate for the current task. Finally, the task’s textual instruction and visual inputs are routed to the selected policy for inference, which outputs actions to execute the task. 

This routing requires only execution records of candidate policies on some tasks and does not require further training on these data, \ie a training-free scheme. Robotic manipulation is a complex problem domain involving multimodal inputs and causal reasoning; building a neural-network-based router that effectively models policy routing in an open task space incurs high data-collection and training costs and, given rapid iteration in robotics, faces continual-learning challenges as new policies enter the pool~\cite{wang2025icl}. In contrast, our routing introduces limited latency. Experiments (\cref{tab:time_exp}) show it is insignificant relative to overall execution time.

\subsection{New-Policy Onboarding}

\PAR{Evaluation task set.}
When a new policy is onboarded into the pool, first determine on which tasks to evaluate it. For a single-task policy, evaluate only on its corresponding training task. For language-conditioned multi-task policies, such as VLA or interface orchestration~\cite{liang2023code}, cluster all seen tasks in the system and test only on the representative task of each cluster. Clustering is based on the task representation (\cref{retriever}), \ie a multimodal embedding for representing a task, with a hyperparameter controlling the number of clusters. We observed over a substantial number of experiments that, on tasks with similar task representations, the performance ranking of policies in the pool remains relatively stable.

\PAR{Post-execution processing.}
Evaluator first applies scripts and task-specific Python utility functions to process video, producing video metadata and selected frames, then observes and summarizes execution as video summary. Researchers typically watch a policy executing a task and, with only a small number of trials, can gain coarse insight into the model’s actual capability; Evaluator simulates this process. Recorder stores the video metadata and video summary produced by Evaluator in Database as new records indexed by the multimodal task representation, and updates Router context.

\PAR{Main advantages.}
(1) Scalability: supports heterogeneous new policies of arbitrary structure and is training-free. (2) Efficiency: Evaluator enables onboarding with few evaluations to obtain sufficient information; clustering by task representation requires testing only on representative tasks per cluster rather than all known tasks, which is meaningful under rapidly emerging robotic manipulation work.

\subsection{Online Feedback}

A process running in parallel with the inference pipeline enables optimization at each trial. Processing is identical to onboarding. For each online execution, Evaluator extracts video metadata and selected frames, then summarizes task execution as video summary. Recorder adds Evaluator outputs to Database and updates Router context.

Two optional human-in-the-loop procedures allow researcher participation in evaluation. Based on different readings of the execution video, a researcher can augment or revise the video summary produced by Evaluator. A researcher can also request another run on the same task with a suboptimal policy to observe and compare performance, then provide feedback. Experiments show that researcher involvement can correct or refine Evaluator judgment of policy execution to some extent.

Because this process runs in a parallel thread, online feedback does not add time overhead to each online inference. Routing decisions in RoboRouter originate from two forms of memory: historical execution records in Database and context maintained by Router. Distribution shifts in factors such as environment or object appearance may weaken these memories. Online feedback provides a path to continuously use the latest experience to calibrate and optimize routing decisions. Human-in-the-loop procedures provide a way to leverage researcher prior knowledge for additional optimization.

\section{Experiments}
\label{sec:Experiments}

We conduct a comprehensive experimental evaluation of RoboRouter and seek to answer the following research questions: (1) How does RoboRouter perform on robotic manipulation tasks across simulation benchmarks and real-world deployments? (2) How well does RoboRouter function as a router? (3) To what extent do system configurations, including modules and hyperparameters, contribute to overall performance?

\subsection{Experimental Setup}

\begin{figure}[t]
  \centering
  \includegraphics[width=\linewidth]{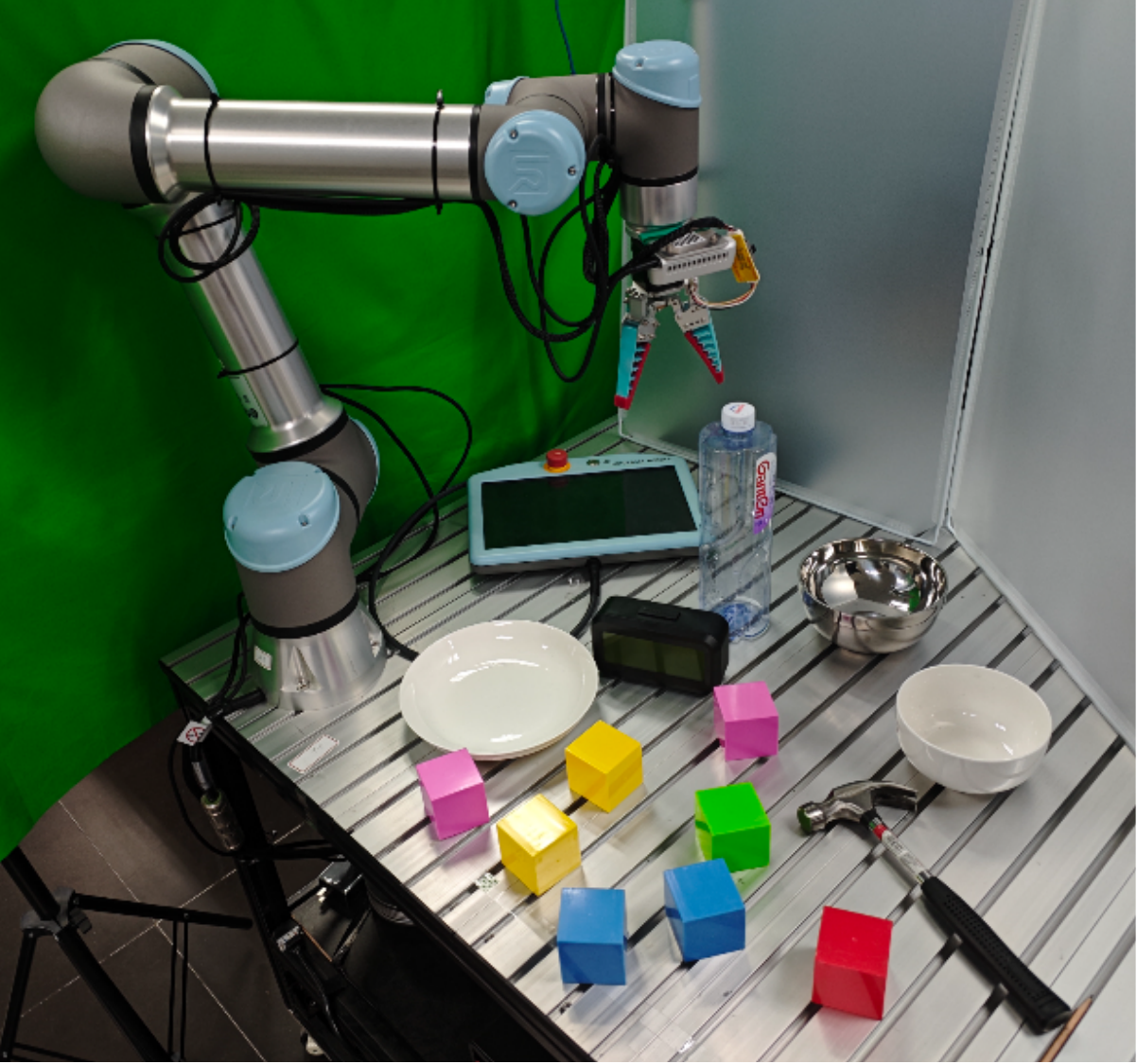}
  \caption{Real-world experimental setup. A desktop platform with a UR5e 6-DoF arm, a motor-driven parallel gripper, and a single Intel RealSense D435i RGB-D camera, together with task-related objects.}
  \label{fig:hardware}
\end{figure}

\PAR{Simulation benchmarks.}
We use RoboTwin 2.0~\cite{chen2025robotwin} as the simulation benchmark to evaluate our approach. RoboTwin 2.0~\cite{chen2025robotwin} is a data generation and benchmarking platform for bimanual manipulation. It builds on the RoboTwin-OD object library with 731 instances across 147 categories, instantiates 50 bimanual tasks, and applies structured domain randomization along five axes.

\PAR{Real-World settings.}
As shown in~\cref{fig:hardware}, we deploy a desktop platform with a UR5e 6-DoF arm and a stepper-motor-driven 1-DoF gripper. Control and communication use URScript with RTDE python interface. Perception uses a single Intel RealSense D435i RGB-D camera. Our computations run on an Intel i7-14700KF CPU with two NVIDIA RTX A6000 GPUs.

\PAR{Baselines.}
Baselines cover three families: single-task policies~\cite{zhao2023learning,chi2025diffusion,ze20243d}, multi-task VLA models~\cite{liu2024rdt,black2024pi_0}, and interface orchestration based on LLM-generated code~\cite{liang2023code}. Only a subset provides public checkpoints on RoboTwin 2.0; we reproduce the remaining checkpoints to match success rates reported in related work. Hyperparameters strongly affect performance; we follow the settings described in RoboTwin 2.0 and align training and inference settings whenever possible.

\PAR{Evaluation metrics.}
For system-level evaluation, we report task-completion success rate and execution time. For routing, we measure consistency with ground-truth routing and routing latency.

\PAR{Implementation details.}
We use Ops-MM-embedding-v1-2B as the multimodal embedding model and Qdrant as the vector database. For reranking, we adopt Qwen3-VL-2B-Instruct~\cite{yang2025qwen3} as the multimodal re-ranker. 
We adopt a logit-based batch reranking scheme that scores all retrieved candidates in a single forward pass using yes/no token probabilities.
We select GPT-4o~\cite{hurst2024gpt} as the primary vision-language model for building our multi-agent system. In the ablation study, we compare against commonly used VLMs (Gemini 1.5 Pro~\cite{team2024gemini}, GPT-4o mini).

\subsection{Main Results}

Results show RoboRouter work across diverse settings.

\PAR{Simulation results.}
In RoboTwin 2.0, we select 20 representative tasks under the simple mode and run 100 trials per task. RoboRouter builds a policy pool from five baselines and can benefit when policy pool is augmented with stronger policies on certain tasks. Note that ACT, DP, and DP3 are single-task policies that require a separate checkpoint for each task. As shown in \cref{tab:sim_exp}, RoboRouter attains 79.9\% mean success, outperforming every individual policy and, on a substantial subset of tasks, slightly surpassing the best policy on that task. Trials within a task include domain randomization such as changes in object pose, creating a finer-grained distinction than task category. By constructing a multimodal task representation, RoboRouter captures configuration details that correlate with policy performance and routes each incoming episode not only to the policy best for that category but to the policy that has performed best under nearly the same configuration (\eg object placement). This yields state-of-the-art performance within single task categories.

\begin{table*}[t]
  \centering
  \caption{Success rates on 20 representative tasks from the RoboTwin 2.0 benchmark (easy-mode evaluation). While individual baselines occasionally dominate on specific tasks, RoboRouter leverages the policy pool to achieve strong performance across most tasks and delivers the highest average success rate}
  \label{tab:sim_exp}
  \small
  \begin{tabular}{@{}lccccccc@{}}
    \toprule
    \textbf{Simulation Tasks} & \textbf{ACT} & \textbf{DP} & \textbf{DP3} & \textbf{RDT} & $\bm{\pi}_0$ & \textbf{Code as Policies}& \textbf{RoboRouter(Ours)} \\
    \midrule
    Adjust Bottle            & \bestbg{98\%} & 91\% & 95\% & 76\% & 85\% & 64\% & \bestbg{98\%} \\
    Beat Block Hammer        & 50\% & 40\% & 74\% & 78\% & 39\% & 53\% & \bestbg{81\%} \\
    Click Alarmclock         & 27\% & 55\% & 74\% & 58\% & 60\% & \bestbg{77\%} & 76\%  \\
    Click Bell               & 57\% & 52\% & 78\% & \bestbg{81\%} & 45\% & 54\% & 79\% \\
    Dump Bin Bigbin          & 67\% & 47\% & 82\% & 66\% & 84\% & 49\% & \bestbg{85\%} \\
    Grab Roller              & 90\% & \bestbg{96\%} & 95\% & 73\% & 94\% & 79\% & \bestbg{96\%} \\
    Handover Mic             & 81\% & 51\% & 94\% & 89\% & \bestbg{96\%} & 43\% & \bestbg{96\%} \\
    Lift Pot                 & 86\% & 37\% & 82\% & 71\% & 82\% & 83\% & \bestbg{87\%} \\
    Move Can Pot             & 16\% & 34\% & 65\% & 21\% & 53\% & 44\% & \bestbg{71\%} \\
    Move Playingcard Away    & 30\% & 42\% & 63\% & 39\% & 48\% & 39\% & \bestbg{65\%} \\
    Open Laptop              & 57\% & 51\% & 84\% & 62\% & \bestbg{87\%} & 60\% & 86\% \\
    Pick Dual Bottles        & 32\% & 26\% & 62\% & 45\% & 59\% & 17\% & \bestbg{63\%} \\
    Place Burger Fries       & 50\% & \bestbg{74\%} & 70\% & 53\% & 71\% & 72\% & \bestbg{74\%} \\
    Place Container Plate    & 70\% & 39\% & 85\% & 78\% & 87\% & 51\% & \bestbg{89\%} \\
    Place Empty Cup          & 60\% & 35\% & 57\% & 56\% & 36\% & 58\% & \bestbg{67\%} \\
    Put Bottles Dustbin      & 26\% & 20\% & 59\% & 21\% & 53\% & 55\% & \bestbg{61\%} \\
    Put Object Cabinet       & 14\% & 40\% & \bestbg{71\%} & 33\% & 67\% & 21\% & 69\% \\
    Shake Bottle             & 75\% & 65\% & 97\% & 76\% & \bestbg{98\%} & 54\% & \bestbg{98\%} \\
    Stack Bowls Three        & 49\% & 64\% & 58\% & 53\% & 62\% & 52\% & \bestbg{65\%} \\
    Stack Bowls Two          & 83\% & 62\% & 84\% & 78\% & \bestbg{92\%} & 64\% & 91\% \\
    \midrule
    Average                  & 55.90\% & 51.05\% & 76.45\% & 60.35\% & 69.90\% & 54.45\% & \bestbg{79.85\%} \\
    \bottomrule
  \end{tabular}
\end{table*}

\PAR{Real-world validation.}
We evaluate five representative real-world tasks with 20 repeated trials per task. RoboRouter builds a policy pool comprising four comparison baselines. We apply domain randomization, including multiple items within a category and varied object placements. As shown in \cref{tab:real_exp}, across real-world tests, RoboRouter achieves significantly higher performance than any single baseline policy.

\begin{table}
  \centering
  \caption{Real-world results on five representative tasks, shown as success counts over 20 episodes.}
  \label{tab:real_exp}
  \small
  \renewcommand{\arraystretch}{1.1}
  \begin{tabular}{@{}lccccc@{}}
    \toprule
    \textbf{Tasks} & \textbf{ACT} & \textbf{DP} & \textbf{RDT} & $\bm{\pi}_0$ & \textbf{Ours} \\
    \midrule
    Adjust Bottle         & \bestbg{8/20} & 5/20 & 2/20 & 4/20 & 7/20 \\
    Beat Block Hammer     & 4/20 & 2/20 & \bestbg{9/20} & 2/20 & \bestbg{9/20} \\
    Click Alarmclock      & 2/20 & 7/20 & 8/20 & 8/20 & \bestbg{10/20} \\
    Open Laptop           & 4/20 & 2/20 & 4/20 & 8/20 & \bestbg{9/20} \\
    Place Container Plate & 8/20 & 2/20 & 10/20 & \bestbg{12/20} & \bestbg{12/20} \\
    \midrule
    Average               & 26\% & 18\% & 33\% & 34\% & \bestbg{47\%} \\
    \bottomrule
  \end{tabular}
\end{table}

\PAR{Average per-trial execution time.}
Using a evaluation script across simulation and real-world settings, we report mean time per trial (\cref{tab:time_exp}). RoboRouter introduces modest overhead, yet differences remain small, which is expected since router components add latency. Given pronounced gains in success rate, this increase is clearly acceptable. Note that, under our protocol, higher success directly lowers measured mean time: simulation benchmark applies a per-trial timeout to terminate obvious failures, typically longer than a successful executions to avoid prematurely stopping runs that might still succeed. Consequently, fewer failures yield lower average time. In real-world evaluation, failures end at a maximum duration or via task-specific triggers (for example, in beat block hammer, knocking the hammer over).

{
\setlength{\tabcolsep}{5pt}
\begin{table}
  \centering
  \caption{Average per-trial execution duration (s) across tasks.}
  \label{tab:time_exp}
  \small
  \renewcommand{\arraystretch}{1.1}
  \begin{tabular}{@{}lccccccc@{}}
    \toprule
    \textbf{Settings} & \textbf{ACT} & \textbf{DP} & \textbf{DP3} & \textbf{RDT} & $\bm{\pi}_0$ & \textbf{CaP} & \textbf{Ours} \\
    \midrule
    RoboTwin 2.0 & 34 & \worstbg{39} & 27 & 33 & 31 & 37 & 38 \\
    Real-World   & 19 & 24 & -  & 20 & 21 & -  & \worstbg{26} \\
    \bottomrule
  \end{tabular}
  \vspace{-1em}
\end{table}
}

\PAR{Routing capability evaluation.}
Routing Accuracy measures overall correctness by counting consistency between routing choices and ground truth. Here, ground truth denotes the policy with highest success rate under fixed task configuration, including instruction and all scene content. As shown in \cref{tab:route_exp}, RoboRouter delivers strong Routing Accuracy, indicating effective memory use to dispatch manipulation tasks to the most suitable policy. Routing Latency quantifies extra delay introduced by routing components during each inference, outside policy execution. Results show RoboRouter sustains high accuracy with latency acceptable in most scenarios.

\begin{table}
  \centering
  \caption{Routing accuracy and routing latency across RoboTwin 2.0 and Real-World settings.} 
  \label{tab:route_exp}
  \small
  \renewcommand{\arraystretch}{1.1}
  \begin{tabular}{@{}lcc@{}}
    \toprule
    \textbf{Setting} & \textbf{Accuracy (\%)} & \textbf{Latency (s)  } \\
    \midrule
    RoboTwin 2.0 & 96.32 & 4.79 \\
    Real-World   & 98.63 & 4.54 \\
    \bottomrule
  \end{tabular}
\end{table}

\subsection{Ablation Studies}

RoboRouter comprises many components, we design an ablation suite to quantify each component’s contribution to overall performance.

Without multimodal retrieval, instruction-only retrieval yields clear degradation. Factors that influence policy behavior, such as object pose and scene layout, are absent from instructions, so retrieval returns records with similar wording yet different physical conditions. Multimodal task representation encodes these cues, enabling retrieval of trials that match performance-relevant factors and guiding routing more effectively. Removing Recorder produces a small drop. Recorder manages updates to Router context and historical execution database; its removal disables dynamic refresh of Router context while keeping only simple appends to historical execution database. Router context summarizes per-policy performance across task clusters and reduces spurious shifts in the recorded distribution caused by chance events during task execution, contributing modest gains.

\begin{table}[t]
  \centering
  \caption{Ablations on component variants, reported as average success rate (\%) in RoboTwin 2.0 and real-world settings.}
  \label{tab:abla_exp}
  \small
  \renewcommand{\arraystretch}{1.1}
  \begin{tabular}{@{}lcc@{}}
    \toprule
    \textbf{Ablation Setting} & \textbf{RoboTwin 2.0} & \textbf{Real-World} \\
    \midrule
    w/o MM Retriever & 78.40\% & 43\% \\
    w/o Recorder          & 78.75\% & 46\% \\
    w/o Evaluator       & 69.80\% & 37\% \\
    w/o Online Feedback                    & 78.15\% & 44\% \\
    VLM: Gemini-1.5-Pro                    & 76.55\% & 45\% \\
    VLM: GPT-4o-mini                       & 76.60\% & 44\% \\
    Ours           & \bestbg{79.90\%} & \bestbg{47\%} \\
    \bottomrule
  \end{tabular}
  \vspace{-1em}
\end{table}

\begin{figure*}[t]
  \centering
  \includegraphics[width=\textwidth]{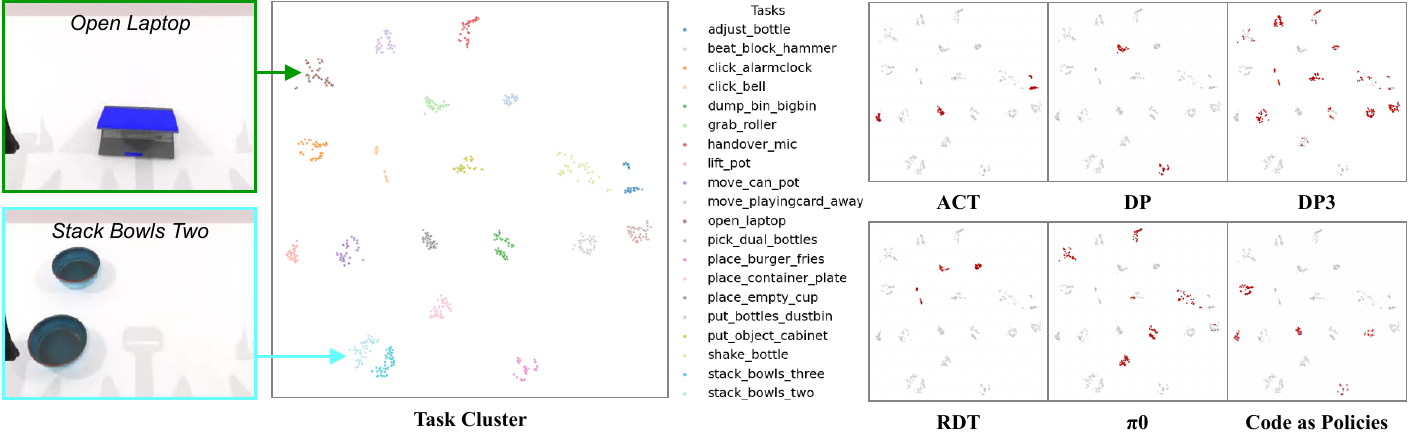}
  \caption{t-SNE Projection of Multimodal Task Representations from 20 RoboTwin 2.0 Tasks. Colored by Task Category and Highlight Routing Selected Policies.}
  \label{fig:tsne}
  \vspace{-0.5em}
\end{figure*}

Removing Evaluator causes a substantial drop, since system only counts per-cluster success rates without per-episode analysis. Disabling online feedback eliminates learning from fresh executions and yields a minor decline. For the VLM choice, GPT-4o serves as primary backend; substituting lighter VLMs such as Gemini 1.5 Pro or GPT-4o mini leads to slight degradation, suggesting nontrivial dependence on VLM capability. Further fine-grained ablation studies are discussed in supplementary material.

\subsection{Analysis and Discussion}

In RoboTwin 2.0 we evaluate 20 representative manipulation tasks (\cref{tab:abla_exp}). Extensive executions on these tasks yield a t-SNE visualization (\cref{fig:tsne}), where each point corresponds to a task representation (\cref{retriever}) of a task configuration as defined. Points from same task cluster tightly, while trials from one task but with different domain randomization, such as instruction wording or object pose, appear as nearby yet distinct points. This structure indicates that task representation capture fine-grained differences across task instances and support retrieval execution records of highly relevant task configuration for routing. Routing outcomes are also visualized (\cref{fig:tsne}). Resulting routing pattern aligns with our accumulated experience in robotic manipulation and with reports in related work, which supports effectiveness of routing mechanism.

We further analyze failure modes. Policies that fail frequently in early stages of an episode tend to have low overall success rate. On beat block hammer, for example, a weaker policy often fails by never grasping the hammer, whereas a stronger policy mainly fails in later stages when the hammer stops just above the block, consistent with common observations in manipulation research. Failure-mode statistics therefore correlate strongly with task-level performance, and inspection of a small number of trials already provides a reasonable estimate of policy competence. Evaluator emulates human researchers reviewing execution videos and acquires this ability .

\section{Conclusion}
In this work, we propose RoboRouter, a training-free, multi-agent policy routing framework for robotic manipulation, constitutes a novel paradigm in this domain. For each incoming task, the system constructs a task representation, retrieves relevant historical execution records from a historical-execution database, infers the policy in the pool best-suited, and executes it directly. It then generates structured feedback from the execution, writes it back to the database, and updates the routing context, forming an online closed loop. New policies can integrated with only lightweight evaluation. As a simple yet effective architecture, RoboRouter offers three key advantages: (i) routing construction and new-policy integration are training-free; (ii) it is agnostic to policy internals and can integrate with any off-the-shelf manipulation policy, fully leveraging community resources; and (iii) it continually learns and calibrates from online runtime feedback. Experiments across multiple settings on RoboTwin simulation benchmark and real-world platform show that RoboRouter significantly improves overall performance, validating the feasibility of this paradigm for robotic manipulation. RoboRouter provides new insights toward building open, scalable robotic manipulation systems.
\clearpage
\appendix

\section*{Appendix}
\label{sec:Appendix}

\section{Additional Experimental Settings and Ablation Study}

\subsection{More Details about Experimental Setup}

For the Code as Policies~\cite{liang2023code} baseline, we implement a simple but faithful reproduction of the original method. We first expose a set of task-relevant control and perception interfaces in either the RoboTwin 2.0 simulation environment~\cite{chen2025robotwin} or the real-world setup, including end-effector pose control, gripper open/close control, object state queries (implemented via either vision foundation models~\cite{ravi2024sam} or privileged simulator information) and so on. Given these interfaces and a natural language instruction, we then prompt LLM to synthesize executable policy code that calls provided interfaces to complete manipulation task.

In ablation studies in which we remove our multimodal retrieval component and restrict the system to text-only retrieval, we adopt LightRAG~\cite{guo2024lightrag} as a widely used retrieval-augmented generation framework for text documents.
We design Evaluator to mimic a human researcher: instead of only logging binary success or failure, it analyzes short video snippets of each rollout to extract richer signals about how and why a policy succeeds or fails. This allows system to obtain substantially more information from each evaluation episode and, therefore, to reduce the total number of episodes required to characterize a new policy on a given task. Concretely, when registering a new policy into policy pool, the number of evaluation trials per task is treated as a hyperparameter. Empirically, we find that using 10 trials per task is sufficient in our setting. Beyond roughly 10 trials, evaluation pipeline rarely discovers qualitatively new information; additional rollouts primarily refine empirical success rates and other statistics. In contrast, many standard simulation benchmarks evaluate a policy with on the order of 100 episodes per task for a full evaluation, so using only 10 episodes at registration time highlights the efficiency gains provided by Evaluator.

\subsection{Additional Ablation Study}

\begin{table}
  \centering
  \caption{Ablations on Evaluator variants, reported as average success rate (\%) in RoboTwin 2.0~\cite{chen2025robotwin} and real-world settings. Best results are bold and shaded in blue, and second-best results are underlined and shaded in gray.}
  \label{tab:abla_evaluator}
  \small
  \renewcommand{\arraystretch}{1.1}
  \begin{tabular}{@{}lcc@{}}
    \toprule
    \textbf{Ablation Setting} & \textbf{RoboTwin 2.0} & \textbf{Real-World} \\
    \midrule
    0 intermediate frames              & 79.25\%            & \bestbg{48\%} \\
    2 intermediate frames (ours)       & \secondbg{79.90\%} & \secondbg{47\%} \\
    4 intermediate frames              & 79.80\%            & 46\% \\
    6 intermediate frames              & \bestbg{79.95\%}   & 46\% \\
    w/o Python tools                   & 77.70\%            & 41\% \\
    w/o VQA                            & 76.35\%            & 42\% \\
    \bottomrule
  \end{tabular}
\end{table}

\begin{figure*}
  \centering
  \includegraphics[width=0.95\linewidth]{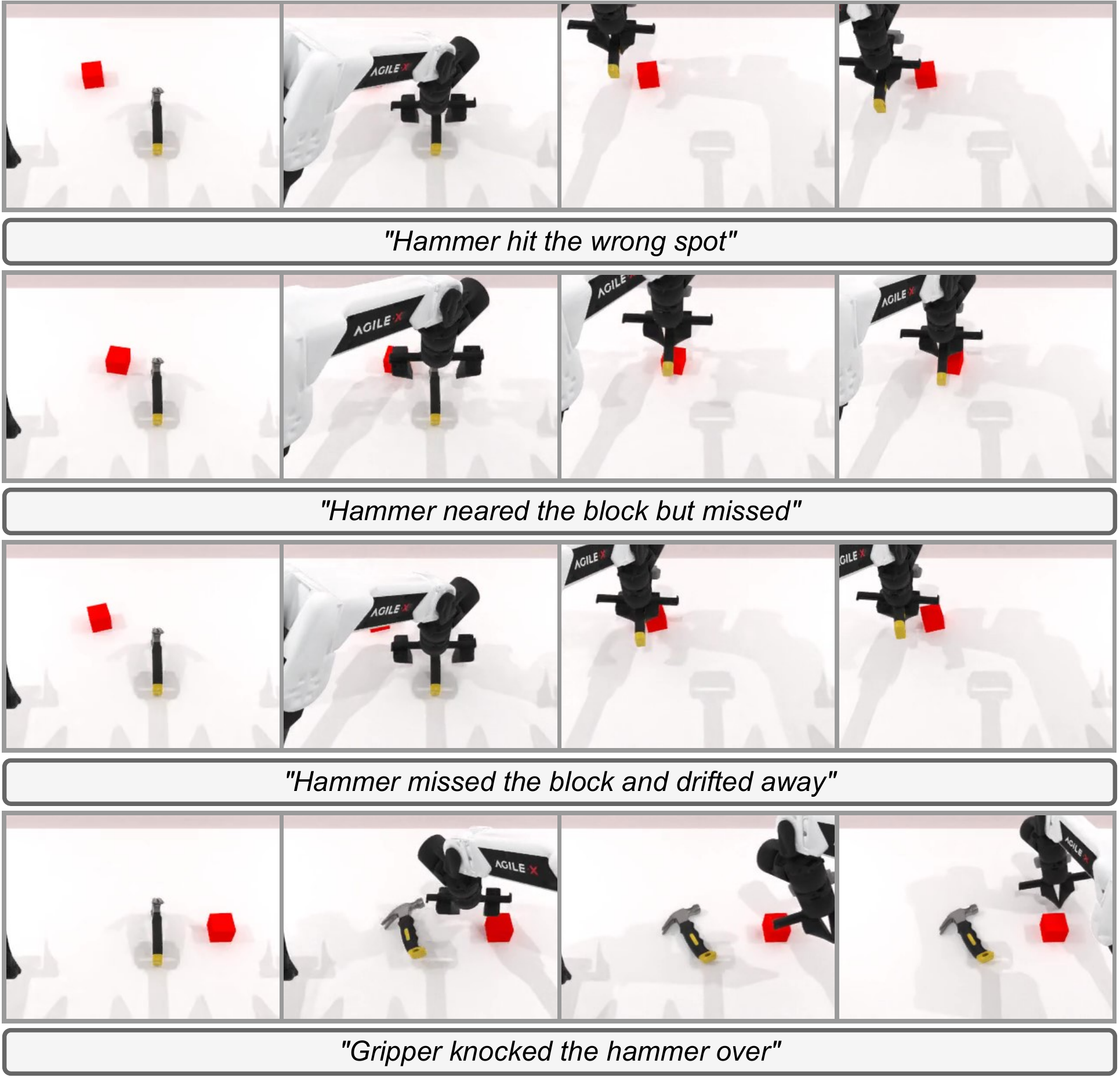}
  \caption{Examples of VQA outputs that correctly summarize representative failure cases under the 2-intermediate-frame setting.}
  \label{fig:vqafail}
\end{figure*}

We conduct additional ablation studies to better understand the importance of several fine-grained design choices in Evaluator.
The first question concerns how the number of video keyframes affects the overall system performance.
Following observations from related work~\cite{li2025cogvla} suggesting that the first and last frames of a trajectory often capture most of the task-relevant information, we include the initial and final observations and vary only the number of intermediate frames, which are uniformly sampled from the rollout. We evaluate configurations with 0, 2, 4, and 6 intermediate frames. As shown in \cref{tab:abla_evaluator}, the number of intermediate frames has only a mild impact on performance, indicating that Evaluator, and consequently RoboRouter, can already operate effectively with very sparse visual summaries of each trajectory. Under the 2-intermediate-frame setting, VQA module can successfully capture and summarize several representative failure cases, as illustrated in \cref{fig:vqafail}.

A second question is how the two complementary information channels inside Evaluator, namely the VQA module and the task-specific Python tool functions, contribute to overall performance. Intuitively, VQA provides flexible and often high-level descriptions of what happened in the rollout, while the Python tools implement precise but narrow measurements (\eg ``did the hammer touch the block?'') derived from task-specific signals. To disentangle their roles, we consider two ablations: w/o tools, where Evaluator relies solely on VQA outputs without any Python tool functions, and w/o VQA, where Evaluator has access only to the Python tools but no VQA descriptions. As summarized in \cref{tab:abla_evaluator}, both channels make clear and complementary contributions: removing either VQA or the Python tools leads to a noticeable drop in performance, confirming that each serves as an important source of execution feedback for Evaluator.

\section{Discussion on System Implementation and Design Choices}

\subsection{Task-Specific Python Tools for Structured Feedback}

Evaluator leverages task-specific Python tool functions to extract structured, rule-based signals from each rollout. These functions operate on low-level observations to produce numerical metrics (\eg distance), which we aggregate as video metadata to complement VQA summaries and to store as part of each execution record. 

Tool functions are authored at task-definition time. In simulation, they are implemented directly on simulator interfaces, such as object poses. In the real world, we approximate the same information using visual foundation models: for example, we employ a pipeline combining DINOv2~\cite{oquab2023dinov2}, SAM~\cite{ravi2024sam}, and k-means clustering to extract 3D semantic keypoints for task-relevant objects, and then compute spatial relations over these keypoints. In Beat Block Hammer task, we identify keypoints corresponding to the top of the block and the hammer head, track their positions over time, and threshold their distance to determine whether contact occurs. The construction of these Python tool functions can be automated via off-the-shelf coding agents, following recent work~\cite{huang2024rekep}.

\subsection{Reranker Design to Mitigate LLM Inductive Biases}

Our reranker is introduced in retriever to mitigate known inductive biases of long-context large language models. Prior studies~\cite{liu2024lost,peysakhovich2023attention,yu2024mitigate} have shown that such models tend to allocate high attention to tokens near the beginning and end of the input context. To address this issue, we apply a lightweight reranker to reorder the candidate records, promoting the most relevant task executions to the front of Router’s input. This design allows Router to process a sufficiently large number of records while effectively focusing its attention on records that are most similar to the current task instance.

\subsection{Task Representation }

Our system uses a multimodal embedding model that takes both images and text as input and outputs an embedding. Our system can directly benefit from future advances in multimodal embedding model. When selecting multimodal embedding models, we consider not only representation quality (\ie performance of the resulting embeddings on downstream tasks), but also inference latency and deployment constraints: some multimodal embedding models are only accessible via remote APIs, which often leads to slower end-to-end response times compared to open-source models that can be deployed locally. 

\begin{figure*}[ht]
  \centering
  \includegraphics[width=0.90\linewidth]{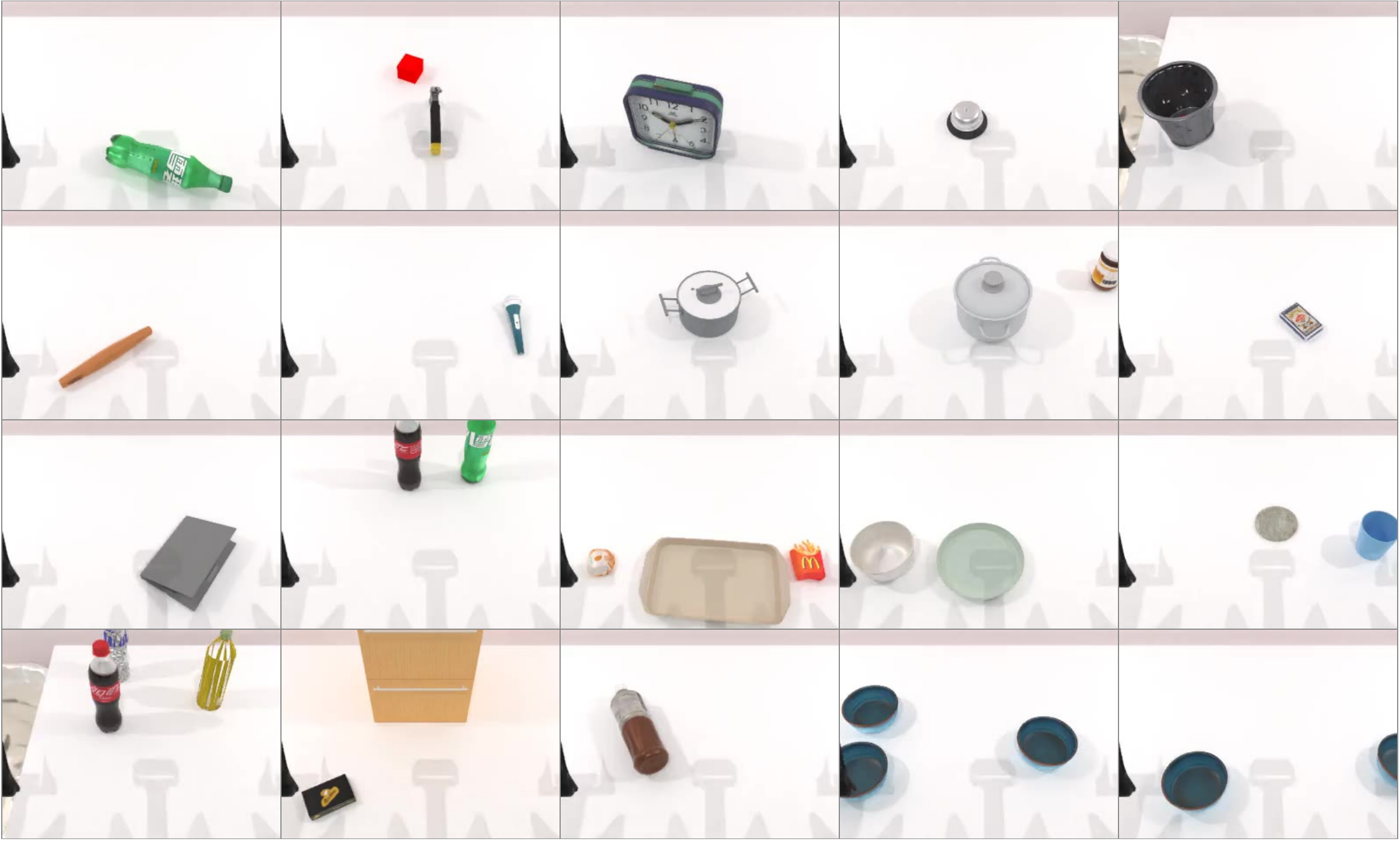}
  \caption{Snapshots of 20 manipulation tasks in RoboTwin 2.0~\cite{chen2025robotwin} simulation benchmark. Experiments on simulation benchmark in main text are based on these tasks}
  \label{fig:20task}
\end{figure*}

\begin{figure*}[ht]
  \centering
  \includegraphics[width=0.90\linewidth]{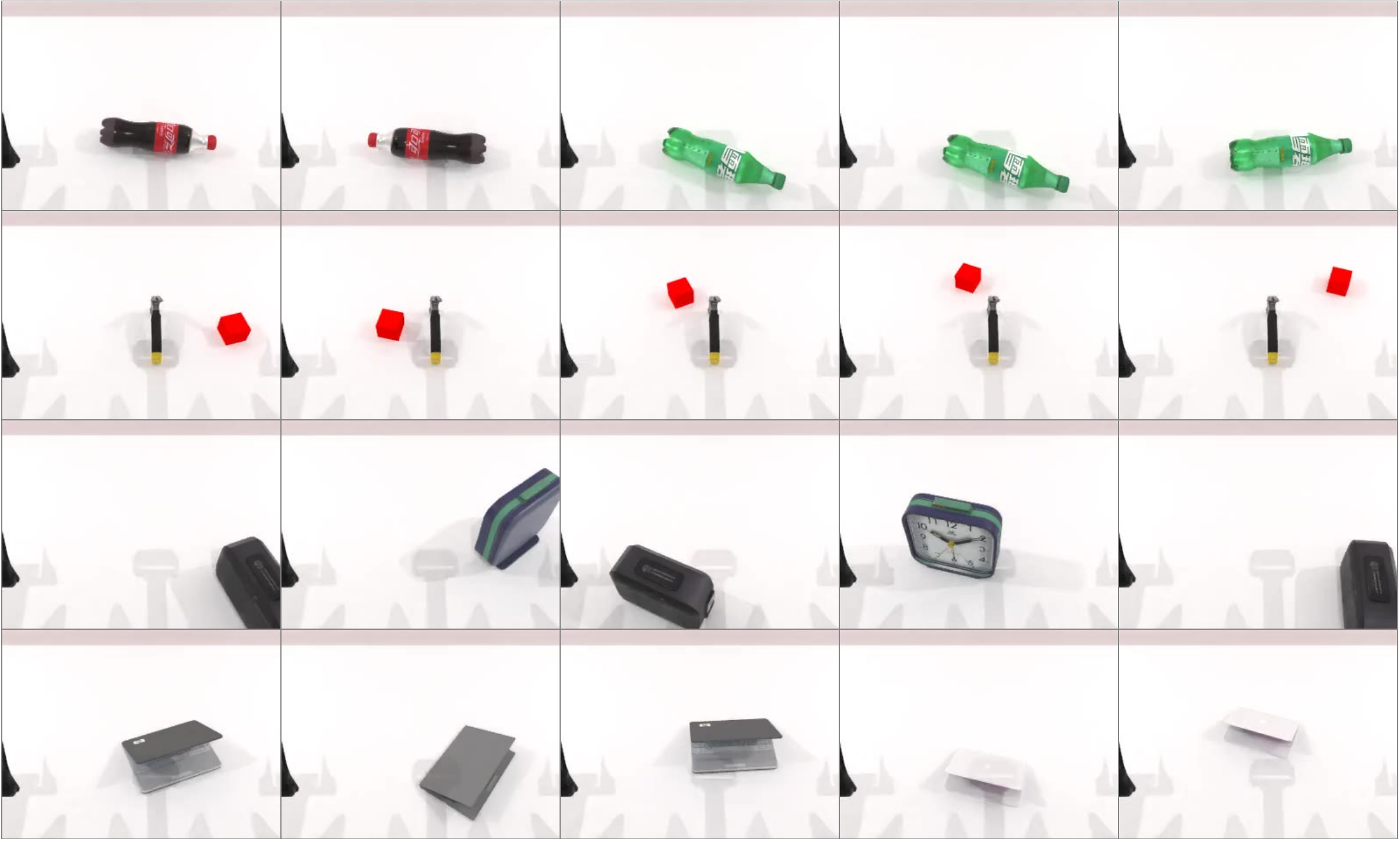}
  \caption{Domain randomization across trials for four example tasks in RoboTwin 2.0~\cite{chen2025robotwin}. For each task we show several rollouts with variations in object instances, poses}
  \label{fig:4random}
\end{figure*}

In RoboTwin 2.0 simulation benchmark~\cite{chen2025robotwin}, visual observations from each trial of the 20 evaluated tasks are fed into multimodal embedding model as a component of the task representation. Some examples are shown in \cref{fig:20task}. Domain randomization across trials (see \cref{fig:4random}), such as changes in object instances, poses, is implicitly encoded into the embedding, enabling the retrieval module to distinguish between task instances that differ only by subtle variations in scene configuration. 

In a real-world setting, many data-driven manipulation policies are sensitive to camera pose~\cite{pang2025learning,mao2025omnid} and degrade noticeably under viewpoint shifts, and different policies may have been trained under different camera placements. At test time, incoming tasks may therefore exhibit novel camera viewpoints. We empirically observe that our multimodal embeddings(task representation) capture such viewpoint information, which allows system to retrieve prior execution records whose camera configuration is similar to the current scene and thereby support more reliable routing decisions.

\section{Discussion of Related Work and Future Works}

Recent years have seen a proliferation of works for robotic manipulation, including world-model-based approaches~\cite{zhou2024dino,pang2025learning} and video-based approaches~\cite{feng2025vidar,tan2025anypos} (generation and parsing). Any such family of manipulation policies can, in principle, be integrated into our framework as additional candidates for routing. 

Currently, our RoboRouter implementation relies on models with a large number of parameters. However, recent progress in agentic reinforcement learning~\cite{zhang2025landscape} and multi-agent reinforcement learning~\cite{sun2024llm} demonstrates that RL-style optimization can substantially improve reasoning ability, suggesting a path toward realizing RoboRouter with much smaller and faster inference models while maintaining decision quality.

Vision-language-action (VLA) model is one of the most promising paradigms. However, under realistic constraints on data scale and model size, researchers have observed that a single VLA often struggles to maintain strong performance over a sufficiently large set of tasks and deployment distributions; achieving broader generalization tends to require sharply increasing data volume, parameters, and training cost in a highly non-linear fashion. Under the RoboRouter perspective, each VLA only needs to be trained on a comparatively narrower distribution where it attains high performance. A diverse pool of VLAs can then be composed via RoboRouter to deliver robust performance over a much broader distribution of tasks, providing more data, training, and parameter efficient alternatives to scaling a single monolithic model.

\clearpage\clearpage
{
    \small
    \bibliographystyle{ieeenat_fullname}
    \bibliography{main}
}

\end{document}